\documentclass[twocolumn]{cinc}
\usepackage{graphicx}
\usepackage{amsmath, amssymb}
\usepackage{booktabs}
\usepackage{multirow}
\begin{document}
\bibliographystyle{cinc}

\title{Physics-Informed Deep Learning for False Ventricular Tachycardia Alarm Reduction in the ICU}

\author {
Athanasios Papastathopoulos-Katsaros$^{1,2}$, Alexandra Stavrianidi$^{3,4}$, Zhandong Liu$^{1,2}$\\
\ \\
$^{1}$ Department of Pediatrics, Baylor College of Medicine, Houston, TX, USA\\
$^{2}$ Jan and Dan Duncan Neurological Research Institute, Texas Children's Hospital, Houston, TX, USA \\
$^{3}$ Institute for Analysis and Numerics, University of M\"unster, Germany \\
$^{4}$ Department of Mathematics, Reed College, Portland, Oregon, USA}
\maketitle
\begin{abstract}
False ventricular tachycardia (VT) alarms are a leading contributor to alarm fatigue in intensive care units. We propose a deep learning framework combining a 1D SE-ResNet with ICU-realistic data augmentations and a physics-informed auxiliary reconstruction task based on the three-element Windkessel hemodynamic model, implemented as a differentiable forward simulation. By requiring the network's latent representation to produce physiologically plausible arterial pressure waveforms, artifact-driven ECG patterns are penalized while true VT remains coherent across modalities. Evaluated on the VTaC benchmark under a strict real-time protocol (10\,s pre-alarm window), our method achieves a Challenge Score of $85.08 \pm 1.65$, a 5-point improvement over prior state-of-the-art. Ablation studies confirm that the physics-informed objective is the primary performance driver, providing gains in accuracy, ${\sim}2\times$ label efficiency, and more localized and clinically meaningful ECG segments.

\end{abstract}

\section{Introduction}

Ventricular tachycardia (VT) is a life-threatening arrhythmia characterized by anomalous ventricular beats exceeding 100\,bpm~\cite{clifford_false_2016}. Because prolonged VT can rapidly lead to sudden cardiac death, ICU monitors are tuned for high sensitivity, making VT alarms among the most prone to false positives~\cite{drew_insights_2014,aboukhalil_reducing_2008}. This contributes to alarm fatigue, a critical patient-safety concern~\cite{fallet_false_2016,chromik_computational_2022}. 

A key difficulty in detecting false alarms is that sensor detachment, patient movement, and electrical interference produce ECG artifacts that closely resemble true arrhythmias, confounding data-driven classifiers. The VTaC benchmark~\cite{lehman_vtac_nodate} provides over 5{,}000 multi-institutional ICU recordings with ECG, photoplethysmography (PLETH), and arterial blood pressure (ABP), enabling evaluation. Prior real-time methods, including supervised CNNs, contrastive models~\cite{contrastive}, and cross-modal VAEs~\cite{wang_biocross_2025}, achieve Challenge Scores up to $80.08$~\cite{lehman2023}. Foundation models~\cite{gu_cardiac_2026} and retrospective
approaches~\cite{farayola_reducing_2025} report higher AUCs,
but the former rely on large external pretraining corpora and the latter use post-alarm information, while still remaining vulnerable to severe sensor noise and previously unseen artifact patterns.

We address this gap by embedding physiological structure into the learning process. Our architecture jointly optimizes classification with cross-modal physics-informed reconstruction of ABP (via a Windkessel simulation) and of PLETH via a data-driven decoder. This constrains the shared representation to support physiologically plausible hemodynamics, thereby penalizing artifact-driven predictions. Our method achieves a Challenge Score of $85.08 \pm 1.65$, a ${\sim}5$-point improvement over prior state-of-the-art, operating strictly within a 10\,s real-time window without external pretraining data.

\section{Methods}
\subsection{Dataset and Preprocessing}
We use the VTaC dataset~\cite{lehman_vtac_nodate}: 5{,}037 expert-annotated alarm events (${\sim}29\%$ true) from three geographically distinct U.S.\ hospitals with different monitors and lead configurations, with the official patient-level 80-10-10 split. Following the real-time protocol, only the final 10\,s before each alarm is used (250\,Hz, $T{=}2500$). Fourteen canonical channels (ECG leads I, II, III, aVR, aVL, aVF, V1--V6, PLETH, ABP) are mapped to fixed slots; unoccupied slots are zero-filled. A binary availability mask $\mathbf{M}\in\{0,1\}^{14\times T}$ is concatenated channel-wise and each channel is independently z-normalized per segment.

\subsection{Architecture}

\begin{figure}[t]
  \centering
  \includegraphics[width=0.7\columnwidth]{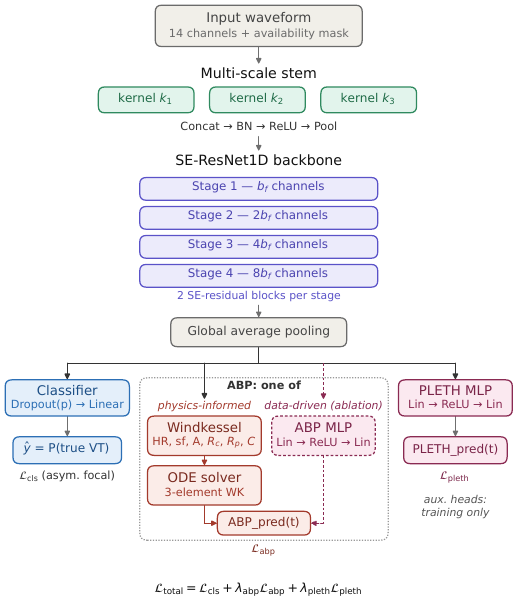}
  \caption{Multi-task architecture. The backbone features are shared
    across a classification head, a physics-informed Windkessel ODE
    reconstruction head for ABP, and a data-driven MLP decoder for
    PLETH.}
  \label{fig:architecture}
\end{figure}

Our backbone is a 1D Squeeze-and-Excitation Residual Network (SE-ResNet1D)~\cite{he_deep_2016,hu_squeeze-and-excitation_2020}. To handle varying heart rates and arrhythmia cycle lengths, we replace the standard initial convolution with a Multi-Scale Stem comprising three parallel 1D convolutional branches (kernel sizes 15, 51, and 201 at stride 2), whose outputs are concatenated, batch-normalized, activated with ReLU, and max-pooled. The core feature extractor consists of four sequential stages, each containing two SE-ResNet blocks (kernel size 7, SE reduction ratio 16), with progressive spatial downsampling and channel doubling at stages 2--4. The output is aggregated via global average pooling, followed by dropout and a linear classification layer. Depending on the inclusion of auxiliary heads, the model contains 2--9M trainable parameters. The multi-task architecture is illustrated in Figure~\ref{fig:architecture}. Hyperparameters were optimized using Optuna's TPE sampler to
maximize the Challenge Score on the validation set \cite{akiba_optuna_2019}. Models were trained with Adam (batch size 64, up to 100 epochs) with early stopping on the validation Challenge Score (patience 20); the decision threshold was selected on the validation set and applied unchanged to the test set. The search included base filter width $\{32,48,64\}$, learning rate
$[10^{-4},10^{-2}]$, weight decay $[10^{-7},10^{-3}]$,
classification dropout $[0.1,0.5]$, the focal-loss penalty,
augmentation intensities, auxiliary-loss weights
$\lambda_{\mathrm{abp}},\lambda_{\mathrm{pleth}}\in[10^{-3},1]$,
and auxiliary target length $\{25,50,100,250,500\}$.

\subsection{Augmentation and Class Imbalance}
We apply five augmentations targeting documented ICU degradation
modes~\cite{iwana_empirical_2021}: temporal jitter ($\pm 50$
samples), additive Gaussian noise, per-channel dropout emulating
sensor detachment~\cite{drew_insights_2014}, amplitude scaling,
and baseline wander (0.1--0.5\,Hz sinusoid).

We address class imbalance through Asymmetric Focal Loss:
\begin{equation}
L_{\text{cls}} = -w_{fn} \cdot y(1{-}\hat{p})^{\gamma^+}\!\log\hat{p} - (1{-}y)\,\hat{p}^{\gamma^-}\!\log(1{-}\hat{p})
\label{eq:loss}
\end{equation}
where $\gamma^{+}=\gamma^{-}=2.0$~\cite{lin_focal_2020} and
$w_{fn}$ is a tunable false-negative weight motivated by the
asymmetric $5\times$ false-negative penalty of the Challenge Score.

\subsection{Physics-Informed Regularization}
To enforce physiological consistency and reduce overfitting to electrical artifacts, we introduce an auxiliary ABP reconstruction task grounded in the three-element Windkessel model. Unlike the PINN paradigm~\cite{raissi_physics-informed_2019}, we do not enforce ODE residuals; instead, we embed a forward physiological model within the supervised model as an auxiliary reconstruction target.

A projection head (64-unit hidden layer, ReLU) predicts six
parameters: heart rate $HR$, systolic fraction $sf$, pulse
amplitude $amp$, proximal resistance $R_c$, peripheral
resistance $R_p$, and compliance $C$, with $HR$ and $sf$ bounded via scaled sigmoids and $amp$, $R_c$, $R_p$, and $C$ constrained to be positive via shifted softplus activations. These parameters drive a differentiable forward simulation in 2 stages:

\emph{Stage 1}: A synthetic ejection proxy models blood flow as a half-sine pulse during systole and zero during diastole:
\begin{equation}
Q(t) = \begin{cases}amp \cdot \sin\!\bigl(\frac{\pi\, t_{\text{mod}}}{sf \cdot T_{\text{cyc}}}\bigr) & t_{\text{mod}} < sf \cdot T_{\text{cyc}} \\ 0 & \text{otherwise}\end{cases}
\end{equation}
where $T_{\text{cyc}} = 60/HR$ and $t_{\text{mod}} = t \bmod T_{\text{cyc}}$.

\emph{Stage 2}: The three-element Windkessel ODE $$C\,dP_{wk}/dt = Q - P_{wk}/R_p$$ is integrated via an exact exponential scheme:
\begin{equation}
P_{wk}[n{+}1] = \gamma\, P_{wk}[n] + R_p(1{-}\gamma)\,Q[n],\;\; \gamma = e^{-\Delta t/(R_pC)}
\end{equation}
which is unconditionally stable for positive $R_p, C$, ensuring well-behaved gradients. The total pressure is $P(t) = R_c Q(t) + P_{wk}(t)$, mapped to the target space via the learnable parameters. The simulation executes in under 3\,ms.

The objective is not to predict ABP; the physics-informed reconstruction acts as \emph{structured regularization} of the shared backbone. Artifact-driven representations are less compatible with coherent hemodynamics and incur higher reconstruction error, propagating corrective gradients through the shared representation.

PLETH is reconstructed via a data-driven MLP decoder (one hidden layer, 128 units, see Figure~\ref{fig:architecture}) leveraging its availability (${\sim}91\%$ of recordings vs.\ ${\sim}36\%$ for ABP). The total training objective is:
\begin{equation}
L_{\text{total}} = L_{\text{cls}} + \lambda_{\text{abp}} L_{\text{abp}} + \lambda_{\text{pleth}} L_{\text{pleth}}
\end{equation}
where $L_{\text{abp}}$ and $L_{\text{pleth}}$ are masked MSE losses computed only when the respective channels are available.

\section{Results}

The primary metric is the PhysioNet 2015 Challenge Score = $(TP + TN)/(TP + TN + FP + 5 \cdot FN)$, which penalizes missed true alarms ($5\times$ weight). All experiments use 5 random seeds; scores are reported as mean $\pm$ SD.

\subsection{Main Results}

Table~\ref{tab:main} compares our models against prior work.

\begin{table}[ht]
\centering
\caption{Performance on VTaC. ``Aug''= augmentations; ``DD''=
data-driven reconstruction; ``Phys''= physics-informed
reconstruction. Challenge score: mean $\pm$ SD over 5 seeds where reported.}
\label{tab:main}
\resizebox{\columnwidth}{!}{%
\begin{tabular}{@{}lccccc@{}}
\toprule
\textbf{Method} & \textbf{Window} & \textbf{Score} & \textbf{AUC} & \textbf{TPR} & \textbf{PPV} \\
\midrule
FCN~\cite{lehman2023}   & 10\,s & $80.08{\pm}2.46$ & .949 & .920 & .717 \\
BioCross~\cite{wang_biocross_2025} & 10\,s & -- & .863 & .814 & .576 \\
CSFM$^\dagger$~\cite{gu_cardiac_2026} & 10\,s & -- & .967 & -- & -- \\
FCNN$^*$~\cite{farayola_reducing_2025} & 6\,min & -- & $\mathbf{.973}$ & .940 & $\mathbf{.950}$ \\
\midrule
Ours: baseline       & 10\,s & $80.84{\pm}2.88$ & .949 & .931 & .708 \\
\;\; + DD Recon      & 10\,s & $81.41{\pm}2.97$ & .956 & .917 & .752 \\
\;\; + Phys Recon    & 10\,s & $82.26{\pm}2.62$ & .947 & .946 & .706 \\
\;\; + Aug           & 10\,s & $81.71{\pm}2.24$ & .953 & .944 & .700 \\
\;\; + Aug \& DD     & 10\,s & $84.16{\pm}0.68$ & .964 & .953 & .732 \\
\;\; + Aug \& Phys   & 10\,s & $\mathbf{85.08{\pm}1.65}$ & .961 & $\mathbf{.958}$ & .742 \\
\bottomrule
\multicolumn{6}{l}{\scriptsize $^*$Retrospective (uses post-alarm data). $^\dagger$Foundation model pretrained on a much larger external corpus.}
\end{tabular}%
}
\end{table}

\begin{figure}[t]
  \centering
  \includegraphics[width=\columnwidth]{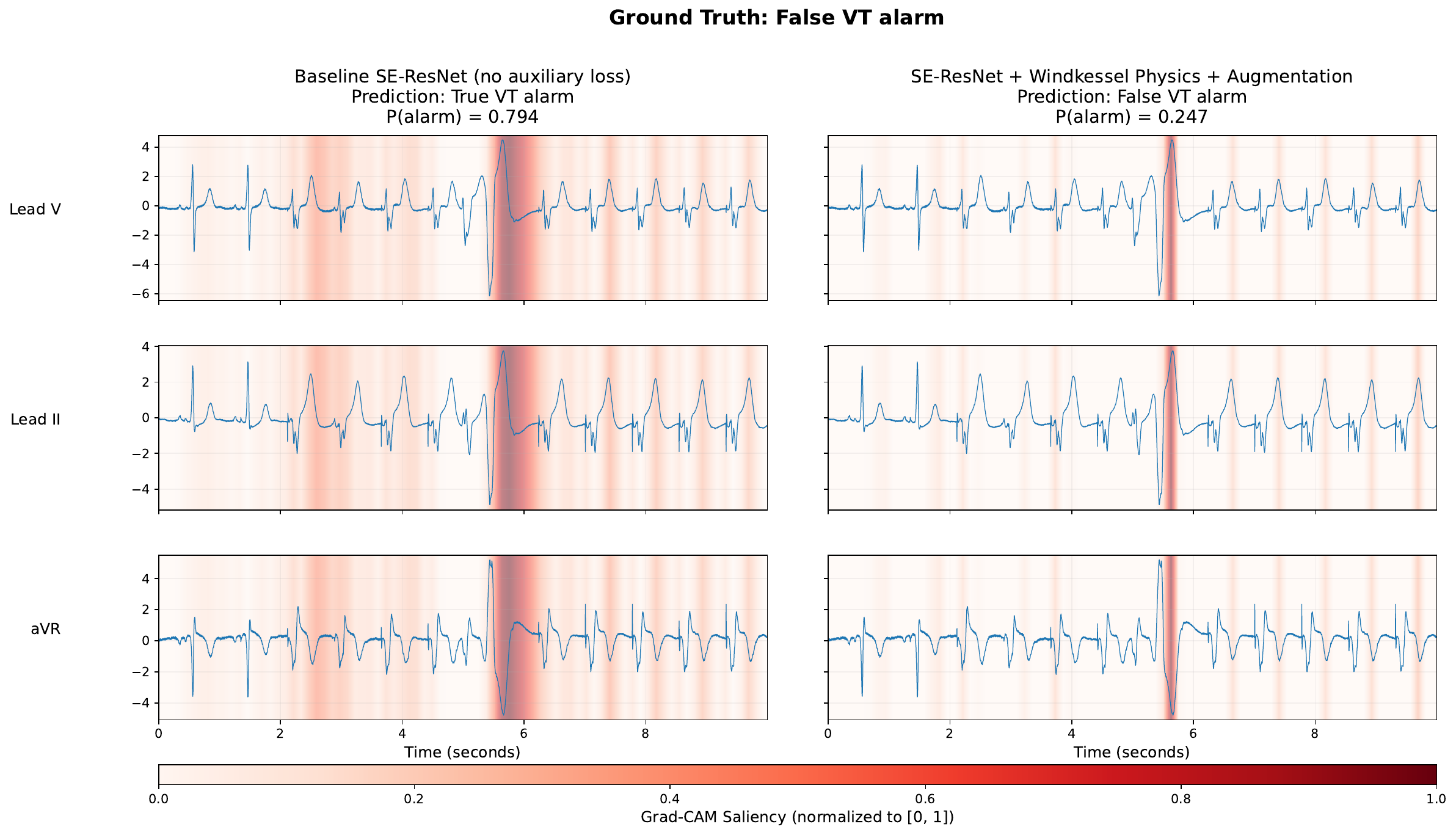}
  \caption{Grad-CAM comparison for a ground-truth false VT alarm. The baseline SE-ResNet (left) is confused by the noisy signal, with diffused saliency across the artifact, leading to an incorrect True VT prediction. In contrast, the physics-informed model (right) has learned to robustly identify non-physiological noise; it strongly localizes the sharp movement artifact and uses this precise detection to correctly reject the false alarm.}
  \label{fig:gradcam}
\end{figure}

\subsection{Ablation Studies}

\emph{Reconstruction paradigm:} Without augmentations, physics-informed reconstruction improves the baseline by $+1.42$ Challenge Score points vs. $+0.57$ for data-driven reconstruction. With augmentations, physics-informed reconstruction improves over augmentation alone by $+3.37$ vs. $+2.45$ for data-driven reconstruction.

\emph{Label efficiency:} Table~\ref{tab:ablation} (top) evaluates all variants at 5\%, 10\%, 20\%, and 50\% label fractions. The physics-informed model at 5\% labels ($64.74\pm2.94$) nearly matches the augmentation-only baseline at 10\% ($65.04 \pm 1.98$), demonstrating approximately $2\times$ label efficiency. It is the only multi-task variant that improves over baseline at every fraction.

\emph{Robustness to missing modalities:} 
Table~\ref{tab:ablation} (bottom) simulates sensor failures on the physics-informed model. Sensor dropout reduces the Challenge Score by $-1.33$ for ABP and $-4.85$ for PLETH. Signal replacement with uncorrelated Gaussian noise (mask retained) is more damaging: $-4.98$ for ABP and $-6.42$ for PLETH. Thus, corrupted signals are more damaging than missing signals.

\emph{Interpretability:} Quantitative localization analysis with Grad-CAM~\cite{selvaraju_grad-cam_2020} was restricted to true-alarm samples ($n=128$), because these contain a genuine VT transition against which temporal localization can be meaningfully assessed. The physics-informed
model produced more localized saliency in 89.1\% of cases, with mean Gini coefficient increasing from $0.52\pm0.11$ to $0.65\pm0.09$. Figure~\ref{fig:gradcam} shows a false alarm misclassified by the baseline but correctly rejected by the physics-informed model.

\begin{table}[ht]
\centering
\caption{(Top) Label efficiency: Challenge Score across label
fractions. All variants include augmentations. (Bottom) Modality robustness on the physics-informed model for a single run.}
\label{tab:ablation}
\resizebox{\columnwidth}{!}{%
\begin{tabular}{@{}lcccc@{}}
\toprule
\multicolumn{5}{c}{\textbf{Label Efficiency (Challenge Score)}} \\
\midrule
\textbf{Variant} & \textbf{5\%} & \textbf{10\%} & \textbf{20\%} & \textbf{50\%} \\
\midrule
Augs baseline (no recon)     & $57.12{\pm}2.79$ & $65.04{\pm}1.98$ & $72.18{\pm}3.24$ & $77.61{\pm}1.91$ \\
PLETH only (DD)         & $62.17{\pm}4.15$ & $68.60{\pm}3.20$ & $70.23{\pm}3.26$ & $76.39{\pm}2.29$ \\
PLETH + ABP (DD)        & $62.72{\pm}3.27$ & $66.85{\pm}3.89$ & $69.63{\pm}2.94$ & $76.47{\pm}2.07$ \\
PLETH + ABP (Phys)      & $\mathbf{64.74{\pm}2.94}$ & $\mathbf{68.90{\pm}4.28}$ & $\mathbf{73.13{\pm}1.36}$ & $\mathbf{78.75{\pm}2.75}$ \\
\midrule
\multicolumn{5}{c}{\textbf{Modality Robustness (Physics Model; Score difference compared to baseline)}} \\
\midrule
\textbf{Condition} & \multicolumn{2}{c}{\textbf{Sensor Dropout}} & \multicolumn{2}{c}{\textbf{Noise Replacement}} \\
\midrule
ABP only   & \multicolumn{2}{c}{$-1.33$} & \multicolumn{2}{c}{$-4.98$} \\
PLETH only & \multicolumn{2}{c}{$-4.85$} & \multicolumn{2}{c}{$-6.42$} \\
Both       & \multicolumn{2}{c}{$-5.18$} & \multicolumn{2}{c}{$-10.15$} \\
\bottomrule
\end{tabular}%
}
\end{table}

\section{Discussion}

The key mechanism underlying the improvement is cross-modal artifact disentanglement. False VT alarms primarily arise from gross signal corruption---electrode detachment, patient movement, electrosurgical interference---that produces wide-complex, high-rate ECG patterns indistinguishable from VT on a single lead~\cite{drew_insights_2014}. The Windkessel reconstruction head penalizes artifact-dependent features because electrical artifacts have no hemodynamic correlate: if the backbone encodes an artifactual ECG as ``VT,'' the simulated pressure waveform will be incoherent and incur high reconstruction loss.

Notably, the ABP reconstruction loss decreases by only approximately 2–-3\% over training, as the lumped Windkessel model can only coarsely approximate the measured waveform. The regularization benefit is disproportionate to the ABP reconstruction accuracy; the value lies in shaping the gradient landscape so that an unconstrained MLP decoder would simply absorb artifacts into flexible weights, while the Windkessel's fixed dynamical structure propagates corrective gradients. Classification requires  $\le$ 4 ms on GPU, and the Windkessel head is removed at inference. However, limitations include the lumped-parameter simplicity of the Windkessel approximation, ABP availability in only 36\% of recordings, and evaluation on a single benchmark. 

\section{Conclusions}
We have shown that embedding a simple hemodynamic forward model as an auxiliary reconstruction task provides a powerful inductive bias for false VT alarm reduction. The physics-informed constraint drives performance gains while improving label efficiency and interpretability, offering a promising strategy for robust clinical classification in high-acuity settings with limited labeled data.

\section*{Code Availability}
Code for training, hyperparameter search and inference is available at \\ \texttt{https://github.com/LiuzLab/vtac-physics}.

\section*{Acknowledgements}
APK and ZL were supported by NIH grant 5R01HG011795, CPRIT
grant RP240131, Chan Zuckerberg Initiative (2023-332162), the Chao Endowment and the Huffington Foundation. AS was supported by the DFG under Germany's Excellence Strategy EXC 2044/2--390685587 and by Reed College startup funds.

\bibliography{sample}

@article{contrastive,
author = {Zhou, Yuerong and others},
year = {2022},
title = {A contrastive learning approach for ICU false arrhythmia alarm reduction},
volume = {12},
journal = {Scientific Reports}
}

@inproceedings{lehman2023, author = {Lehman, Li-wei H. and others}, title = {VTaC: a benchmark dataset of ventricular tachycardia alarms from ICU monitors}, year = {2023}, publisher = {Curran Associates Inc.}, address = {Red Hook, NY, USA}, booktitle = {Proceedings of the 37th International Conference on Neural Information Processing Systems}, articleno = {1686},  location = {New Orleans, LA, USA}, series = {NIPS '23} }

@article{hu_squeeze-and-excitation_2020,
	title = {Squeeze-and-{Excitation} {Networks}},
	volume = {42},
	copyright = {https://ieeexplore.ieee.org/Xplorehelp/downloads/license-information/IEEE.html},
	number = {8},
	urldate = {2026-04-07},
	journal = {IEEE Trans. Pattern Anal. Mach. Intell.},
	author = {Hu, Jie and others},
	year = {2020},
}

@inproceedings{he_deep_2016,
	address = {Las Vegas, NV, USA},
	title = {Deep {Residual} {Learning} for {Image} {Recognition}},
	urldate = {2026-04-07},
	booktitle = {2016 {IEEE} {Conference} on {Computer} {Vision} and {Pattern} {Recognition} ({CVPR})},
	publisher = {IEEE},
	author = {He, Kaiming and others},
	year = {2016},
}

@article{wang_biocross_2025,
	title = {{BioCross}: {A} cross-modal framework for unified representation of multi-modal biosignals with heterogeneous metadata fusion},
	volume = {123},
	shorttitle = {{BioCross}},
	journal = {Information Fusion},
	author = {Wang, Mengxiao and others},
	year = {2025},
}

@article{gu_cardiac_2026,
	title = {Cardiac health assessment across scenarios and devices using a multimodal foundation model pretrained on data from 1.7 million individuals},
	volume = {8},
	number = {2},
	urldate = {2026-04-07},
	journal = {Nat Mach Intell},
	author = {Gu, Xiao and others},
	year = {2026},
}

@inproceedings{farayola_reducing_2025,
	address = {Helsinki, Finland},
	title = {Reducing {False} {Ventricular} {Tachycardia} {Alarms} in {ICU} {Settings}: {A} {Machine} {Learning} {Approach}},
	shorttitle = {Reducing {False} {Ventricular} {Tachycardia} {Alarms} in {ICU} {Settings}},
	booktitle = {2025 10th {International} {Conference} on {Machine} {Learning} {Technologies} ({ICMLT})},
	publisher = {IEEE},
	author = {Farayola, Grace Funmilayo and others},
	year = {2025},
}

@misc{lehman_vtac_nodate,
	title = {{VTaC}: {A} {Benchmark} {Dataset} of {Ventricular} {Tachycardia} {Alarms} from {ICU} {Monitors}},
	shorttitle = {{VTaC}},
	urldate = {2026-04-07},
    year = {2024},
	publisher = {PhysioNet},
	author = {Lehman, Li-wei and others},
}

@inproceedings{akiba_optuna_2019,
	address = {Anchorage AK USA},
	title = {Optuna: {A} {Next}-generation {Hyperparameter} {Optimization} {Framework}},
	shorttitle = {Optuna},
	urldate = {2026-04-07},
	booktitle = {Proceedings of the 25th {ACM} {SIGKDD} {International} {Conference} on {Knowledge} {Discovery} \& {Data} {Mining}},
	publisher = {ACM},
	author = {Akiba, Takuya and others},
	year = {2019},
}

@article{clifford_false_2016,
	title = {False alarm reduction in critical care},
	volume = {37},
	number = {8},
	urldate = {2026-04-08},
	journal = {Physiol. Meas.},
	author = {Clifford, Gari D. and others},
	year = {2016},
}

@article{drew_insights_2014,
	title = {Insights into the {Problem} of {Alarm} {Fatigue} with {Physiologic} {Monitor} {Devices}: {A} {Comprehensive} {Observational} {Study} of {Consecutive} {Intensive} {Care} {Unit} {Patients}},
	volume = {9},
	shorttitle = {Insights into the {Problem} of {Alarm} {Fatigue} with {Physiologic} {Monitor} {Devices}},
	number = {10},
	urldate = {2026-04-08},
	journal = {PLoS ONE},
	author = {Drew, Barbara J. and others},
	year = {2014},
}

@article{aboukhalil_reducing_2008,
	title = {Reducing false alarm rates for critical arrhythmias using the arterial blood pressure waveform},
	volume = {41},
	copyright = {https://www.elsevier.com/tdm/userlicense/1.0/},
	number = {3},
	urldate = {2026-04-08},
	journal = {Journal of Biomedical Informatics},
	author = {Aboukhalil, Anton and others},
	year = {2008},
}

@article{fallet_false_2016,
	title = {False arrhythmia alarms reduction in the intensive care unit: a multimodal approach},
	volume = {37},
	shorttitle = {False arrhythmia alarms reduction in the intensive care unit},
	number = {8},
	urldate = {2026-04-08},
	journal = {Physiol. Meas.},
	author = {Fallet, Sibylle and others},
	year = {2016},
}

@article{chromik_computational_2022,
	title = {Computational approaches to alleviate alarm fatigue in intensive care medicine: {A} systematic literature review},
	volume = {4},
	shorttitle = {Computational approaches to alleviate alarm fatigue in intensive care medicine},
	urldate = {2026-04-08},
	journal = {Front. Digit. Health},
	author = {Chromik, Jonas and others},
	year = {2022},
}

@article{selvaraju_grad-cam_2020,
	title = {Grad-{CAM}: {Visual} {Explanations} from {Deep} {Networks} via {Gradient}-{Based} {Localization}},
	volume = {128},
	shorttitle = {Grad-{CAM}},
	number = {2},
	urldate = {2026-04-09},
	journal = {Int J Comput Vis},
	author = {Selvaraju, Ramprasaath R. and others},
	year = {2020},
}

@article{raissi_physics-informed_2019,
	title = {Physics-informed neural networks: {A} deep learning framework for solving forward and inverse problems involving nonlinear partial differential equations},
	volume = {378},
	shorttitle = {Physics-informed neural networks},
	urldate = {2026-04-09},
	journal = {Journal of Computational Physics},
	author = {Raissi, M. and others},
	year = {2019},
}

@article{lin_focal_2020,
	title = {Focal {Loss} for {Dense} {Object} {Detection}},
	volume = {42},
	copyright = {https://ieeexplore.ieee.org/Xplorehelp/downloads/license-information/IEEE.html},
	number = {2},
	urldate = {2026-04-13},
	journal = {IEEE Trans. Pattern Anal. Mach. Intell.},
	author = {Lin, Tsung-Yi and others},
	year = {2020},
}

@article{iwana_empirical_2021,
	title = {An empirical survey of data augmentation for time series classification with neural networks},
	volume = {16},
	number = {7},
	urldate = {2026-04-09},
	journal = {PLoS ONE},
	author = {Iwana, Brian Kenji and others},
	editor = {Schwenker, Friedhelm},
	year = {2021},
}

\begin{correspondence}
Athanasios Papastathopoulos-Katsaros\\
Department of Pediatrics, Baylor College of Medicine, Houston, TX, 77030, United States of America\\
athanasios.papastathopoulos-katsaros@bcm.edu
\end{correspondence}
\end{document}